\documentclass[US letter, 10pt, conference]{ieeeconf}      

\IEEEoverridecommandlockouts

\usepackage{cite}
\usepackage{amsmath,amssymb,amsfonts}
\usepackage{algorithmic}
\usepackage{graphicx}
\usepackage{textcomp}
\usepackage{xcolor}

\usepackage[pdfstartview=XYZ,
bookmarks=true,
colorlinks=true,
linkcolor=black,
urlcolor=black,
citecolor=black,
bookmarks=true,
linktocpage=true, 
hyperindex=true
]{hyperref}

\usepackage{orcidlink}
   
\begin{document}

© 2022 IEEE. Personal use of this material is permitted.
Permission from IEEE must be obtained for all other uses,
including reprinting/republishing this material for advertising
or promotional purposes, collecting new collected works
for resale or redistribution to servers or lists, or reuse of
any copyrighted component of this work in other works.
This work has been submitted to the IEEE for possible
publication. Copyright may be transferred without notice,
after which this version may no longer be accessible.

Manuscript is accepted to publish in 2022 IEEE-RAS 21st
International Conference on Humanoid Robots (Humanoids).

Please cite the newer, accepted version that has the DOI below: 10.1109/Humanoids53995.2022.10000105

\title{\LARGE \bf
Decentralized Nonlinear Control of Redundant Upper Limb Exoskeleton with Natural Adaptation Law
}

\author{Mahdi Hejrati$^{1}$ \orcidlink{0000-0002-8017-4355}, Jouni Mattila$^{2}$ \orcidlink{0000-0003-1799-4323}
\thanks{*The TITAN (Teaching human-like abilities to heavy mobile manipulators through multisensory presence) project is funded by the Technology Industries of Finland Centennial Foundation and the Jane and Aatos Erkko Foundation Future Makers programme. 2020-2023.}
\thanks{$^{1}$M. Hejrati is with the Faculty of Engineering and Natural Science,
        Tampere University, 7320 Tampere, Finland, Corresponding Author
        {\tt\small mahdi.hejrati at tuni.fi}} %
\thanks{$^{2}$J. Mattila is with the Faculty of Engineering and Natural Science,
        Tampere University, 7320 Tampere, Finland,
        {\tt\small jouni.Mattila at tuni.fi}}%
}


\maketitle
\thispagestyle{empty}
\pagestyle{empty}

\begin{abstract}

The aim of this work is to utilize an adaptive decentralized control method called virtual decomposition control (VDC) to control the orientation and position of the end-effector of a 7 degrees of freedom (DoF) right-hand upper-limb exoskeleton. The prevailing adaptive VDC approach requires tuning of 13\(n\) adaptation gains along with 26\(n\) upper and lower parameter bounds, where \(n\) is the number of rigid bodies. Therefore, utilizing the VDC scheme to control high DoF robots like the 7-DoF upper-limb exoskeleton can be an arduous task. In this paper, a new adaptation function, so-called natural adaptation law (NAL), is employed to eliminate these burdens from VDC, which results in reducing all 13\(n\) gains to one and removing dependency on upper and lower bounds. In doing so, VDC-based dynamic equations are restructured, and inertial parameter vectors are made compatible with NAL. Then, the NAL adaptation function is exploited to design a new adaptive VDC scheme. This novel adaptive VDC approach ensures physical consistency conditions for estimated parameters with no need for upper and lower bounds. Finally, the asymptotic stability of the algorithm is proved with the virtual stability concept and the accompanying function. The experimental results are utilized to demonstrate the excellent performance of the proposed new adaptive VDC scheme.

\end{abstract}

\section{INTRODUCTION}

Generally speaking, exoskeletons are human-robot interfaces (HRIs) that ease interaction and communication between human operators and robots. In the context of teleoperation systems, a human operator needs to have an adequate sense of the remote site and control of the slave robot to accomplish a successful teleoperation task \cite{c1}. As a reasonable solution, an exoskeleton transfers human body movements to the slave robot and provides a sense of the remote site through haptic technology, allowing the operator to execute dexterous tasks.

The aim of our project is to study two upper-limb 7-DoF exoskeletons as master robots and two heavy-duty hydraulic mobile manipulators as slave robots within the concept of multi-master multi-slave (MMMS) teleoperation systems \cite{c2}. Like in our earlier work \cite{c3}, we considered real-world dissimilar master and slave robots, where the range of motion and force amplification was 1:4 for motion and 1:800 for force amplification. This required us to consider the problem of motion/force scaling which resulted complicated stability analysis tackled in \cite{c3}. However, in [3], a commercial 6-DoF desktop haptic manipulator was utilized for controlling a stationary heavy-duty hydraulic slave robot, whereas our current target is to utilize wearable 7-DoF exoskeletons to control multiple mobile slave robots in human-like object manipulation scenarios. Therefore, it is essential to design a control strategy with the capability of handling these highly complex MMMS systems for the required high performance control objectives, such as high transparency as in \cite{c3}. For this ambitious objective, in this paper, decentralized nonlinear control problem of the 7-DoF right-hand upper-limb exoskeleton is examined. In the future, this work will be expanded to the control of MMMS teleoperation systems.

The control problem of one-armed upper-limb exoskeletons has been widely examined. In \cite{c4}-\cite{c5}, adaptive integral terminal sliding mode control (SMC) and integral second-order terminal SMC have been employed to control upper-limb exoskeletons, respectively. In \cite{c6}, an integral second-order terminal SMC with quasi-time delay estimation (QTDE) was employed for force and position control of an upper-limb exoskeleton. Adaptive impedance control with a nonlinear disturbance observer with estimation of human-desired intended motion has been considered in \cite{c7} and \cite{c8}. Other control schemes, such as backstepping \cite{c9}, backstepping-SMC \cite{c10}, and intelligent approaches like fuzzy SMC \cite{c11} have also been applied to the exoskeleton control problem. Although the above-mentioned control schemes have shown good performance, they are not perfect choices for MMMS systems, especially when MMMS teleoperation systems work to manipulate a common object. In such a scenario, the levels of nonlinearity and complexity of the system are intensely high, and the aforementioned approaches can face difficulties in handling them. For example, backstepping can end up with an explosion of complexity, even for a simple system \cite{c12}. Moreover, most of the mentioned works are about the applications of exoskeletons in rehabilitation with one arm, while our goal is to use exoskeletons in MMMS teleoperation systems. However, MMMS systems encompass rehabilitation and telerehabilitation.

The virtual decomposition control (VDC) approach is a decentralized control scheme that breaks the entire complex system into multiple subsystems that are easy to deal with and design control action \cite{c13}. Virtual cutting points (VCPs) are points at which a system is broken down into subsystems. With the introduction of virtual power flows (VPFs) at VCPs as stability connectors, the stability of each subsystem, which is examined distinctly at the subsystem level, is expanded to the whole system and culminates in asymptotic convergence. Such an approach of modulation makes VDC a much more stable control algorithm in a way that altering the control action of one subsystem does not affect the others. Therefore, VDC is able to handle extremely high DoF dissimilar teleoperation systems and perfect choice for our future aims.

So far, the VDC method has been utilized in fields such as hydraulic manipulators \cite{c14}-\cite{c16}, mobile manipulators \cite{c17}, and underwater applications \cite{c18}. Nonetheless, for the exoskeleton application, only \cite{c19} has employed an adaptive VDC scheme to control a 7-DoF exoskeleton with estimation of the joint's moment of inertia and friction. In contrast to \cite{c19}, which examined rehabilitation without motion/force scaling, our aim is to use the exoskeleton in dissimilar MMMS teleoperation systems with high motion/force scaling. However, because adaptive VDC utilizes a projection function \cite{c13} to estimate unknown parameters of the rigid body, it has two major drawbacks: \(i\)) it requires lower and upper bounds of each parameter, and \(ii\)) it designates an adaptation gain to each parameter. On the other hand, the inertial parameter vector of the VDC-based dynamic equation encompasses 13 parameters, which in general, the unique inertial parameter vector must contain 10 parameters \cite{c20}. This means that 13\(n\) adaptation gains and 26\(n\) lower and upper bounds are required for adaptive VDC, where \(n\) is the number of rigid bodies. Therefore, as the DoF of a system increases, like in \cite{c19} and MMMS teleoperation systems, preparing and tuning all these parameters can make VDC time-consuming, despite its benefit of being decentralized.

The aim of this paper is to eliminate the above-mentioned problems by proposing a novel adaptive VDC approach that utilizes NAL \cite{c21} for the estimation of unknown rigid body parameters. The NAL function is defined based on the Riemannian metric and is coordinate invariant with respect to coordinate frames or physical unit selection. This adaptation law only requires one tuning gain for the entire chain of rigid bodies with no need for upper and lower bounds. The contributions of this paper are as follows:

\begin{itemize}

\item Redesigning dynamic equations of VDC into the 10-form inertial parameter vectors.

\item Proposing an asymptotic-stability guaranteed, novel VDC-NAL approach. By doing so, all the 26\(n\) lower and upper bounds are eliminated and 13\(n\) adaptation gains are reduced to 1, while the physical consistency of the estimated parameters is ensured.

\item Applying the presented approach to control a 7-DoF exoskeleton. The commercial exoskeleton ABLE by Haption is used in the experiment to examine the performance of the modified controller.

\end{itemize}

\section{Proposed VDC Structure}

In this section, the prevailing form of VDC is explained and essential theorems and definitions are provided to prove stability of the system in the sense of virtual stability. Then, the dynamic equations of the system are reformulated to suit the presented adaptation law.

\subsection{Virtual Decomposition Control Approach}

Consider \{A\} as a frame that is attached to a rigid body. Then, the 6D linear/angular velocity vector \(^AV\in \Re^6\) and force/moment vector \(^AF\in \Re^6\) can be expressed as follows \cite{c13}:

\begin{equation}\label{equ1}
^AV = [^Av,^A\omega]^T,\quad ^AF = [^Af,\,^Am]^T
\end{equation}
where \(^Av\in \Re^3\) and \(^A\omega\in \Re^3\) are the linear and angular velocities of frame \{A\}, and \(^Af\in \Re^3\) and \(^Am\in \Re^3\) are the force and moment expressed in frame \{A\}, respectively. The transformation matrix that transforms force/moment vectors and velocity vectors between frame \{A\} and \{B\}, where \{B\} is another frame attached to the rigid body, is \cite{c13}

\begin{equation}\label{equ2}
^AU_B = \begin{bmatrix}
^AR_B & \textbf{0}_{3\times3} \\
(^Ar_{AB}\times)\, ^AR_B & ^AR_B
\end{bmatrix}
\end{equation}
where \(^AR_B \in \Re^{3\times3} \) is a rotation matrix between frame \{A\} and \{B\}, (\(\times\)) operator is a skew-symmetric operator defined in \cite{c13}, and \(^Ar_{AB}\) denotes a vector from the origin of frame \{A\} to the origin of frame \{B\}, expressed in \{A\}. Based on the \(^AU_B\), the force/moment and velocity vectors can be transformed between frames as \cite{c13}

\begin{equation}\label{equ3}
^BV =\, ^AU_B^T\,^AV,\quad ^AF =\, ^AU_B\, ^BF.
\end{equation}
Then, the dynamic equation of the free rigid body expressed in frame \{A\} can be derived as follows \cite{c13}:

\begin{equation}\label{equ4}
\Bar{M}_A\frac{d}{dt}(^AV)+\Bar{C}_A(^AV)+\Bar{G}_A=\, ^AF^*
\end{equation}
where \(\Bar{M}_A \in \Re^{6\times6} \) is the mass matrix, \(\Bar{C}_A \in \Re^{6\times6} \) is the centrifugal and coriolis matrix, \(\Bar{G}_A \in \Re^6 \) is the gravity vector, and \(^AF^* \in \Re^6 \) is the net force/moment vector applied to the rigid body. These matrices are \cite{c13}
\begin{equation*}
\Bar{M}_A = \begin{bmatrix}
m_A\, I_3 & -m_A\, (^Ar_{AB}\times) \\
m_A\, (^Ar_{AB}\times) & I_A\, -\, m_A\, (^Ar_{AB}\times)^2
\end{bmatrix}\\
\end{equation*}

\begin{equation*}
\begin{aligned}
\Bar{C}_A &= 
\left[\begin{matrix}
m_A\, (^A\omega\times) \\
m_A\, (^Ar_{AB}\times)\, (^A\omega\times)  &  (^A\omega\times)\, I_A\, +\,
\end{matrix}\right.\\
&
\left.\begin{matrix}
    -m_A\, (^A\omega\times)\, (^Ar_{AB}\times) \\
     I_A\, (^A\omega\times)-m_A\, (^Ar_{AB}\times)\, (^A\omega\times)\, (^Ar_{AB}\times)
\end{matrix}\right]
\end{aligned}
\end{equation*}
\begin{equation}\label{equ5}
\Bar{G}_A = \begin{bmatrix}
m_A\, ^AR_Ig \\
m_A\, (^Ar_{AB}\times)\, ^AR_Ig
\end{bmatrix}
\end{equation}
where \(m_A \in \Re \) denotes the mass of the rigid body, \(I_3 \) is a 3\(\times \)3 identity matrix, \(I_A =\, ^AR_II_o\,^IR_A \) where \(I_o \in \Re^{3\times3} \) denotes the moment of inertia matrix around the center of mass, and \(g = [0,0,9.8]^T \in \Re^3 \) is the gravity vector.

The dynamic equation (\ref{equ4}) can be rewritten as a linear-in-parameter expression:

\begin{equation}\label{equ6}
Y^r_A\theta_A = \Bar{M}_A\frac{d}{dt}(^AV_r)+\Bar{C}_A(^AV_r)+\Bar{G}_A,
\end{equation}
where \(Y^r_A \in \Re^{6\times13} \) is the regressor matrix generated from \(^AV_r\), \(^AV_r\) is the required velocity vector, and \(\theta_A \in \Re^{13}\) is the rigid body inertial parameter vector. Furthermore, the required net force/moment vector can be designed as

\begin{equation}\label{equ7}
^AF^*_r = Y^r_A\hat{\theta}_A + K(^AV_r-^AV),
\end{equation}
where \(\hat{\theta}_A\) is the estimation of the unknown inertial parameter vector, and \(K\) is the symmetric positive-definite matrix. To estimate the inertial parameter vector of each rigid body, the projection function is utilized as follows \cite{c13}:

\begin{equation}\label{equ8}
\Dot{\hat{\theta}}_{Ai}(t) = \rho_is_i\kappa_i,\quad i=1..13
\end{equation}
with 
\begin{equation}\label{equ9}
\kappa_i= \left\{
  \begin{array}{lr} 
      0 & if\; \hat{\theta}_i\leq \underline{\theta}_i\; and\; s_i \leq 0 \\
      0 & if\; \hat{\theta}_i\geq \overline{\theta}_i\; and\; s_i \geq 0 \\
      1 & otherwise
      \end{array}
\right.
\end{equation}
and
\begin{equation}\label{equ10}
s(t) = Y^r_A{}^T(^AV_r-\,^AV)
\end{equation}
where \(\rho_i>0\) is the adaptation gain, and \(\underline{\theta}_i\) and \(\overline{\theta}_i\) are the lower and upper limits of the \(i^{th}\) parameter, respectively. 

Each adjacent subsystem interacts with each other at VCPs with governing dynamics of (\ref{equ4}) and a control law of (\ref{equ7}). Such interactions are designated as VPFs. Considering the fixed frame \{A\}, VPF can be defined as follows \cite{c13}:

\begin{equation}\label{equ11}
p_A = (^AV_r-^AV)^T(^AF_r-\,^AF)
\end{equation}
where \(^AF_r \in \Re^6\) is the required force. VPFs act as stability connectors in the concept of virtual stability and extend the stability of each subsystem to the stability of entire system. The following lemma expresses the virtual stability concept of VDC:

\textbf{Definition 1.}\cite{c13} A single subsystem that is virtually decomposed from complex system with dynamics (\ref{equ4}), control law (\ref{equ7}) with known \(\theta\), and affiliated function \(X(t)\) can be said to be virtually stable if and only if there exists a non-negative accompanying function \(\nu(t)\)

\begin{equation}\label{equ12}
\nu(t) \geq \frac{1}{2}X(t)^TPX(t)
\end{equation}
in a way that

\begin{equation}\label{equ13}
\Dot{\nu}(t) \leq -X(t)^TQX(t)+p_{\underline{A}}-p_{\overline{A}}
\end{equation}
where \(P\) and \(Q\) are two block-diagonal positive-definite matrices, and \(\underline{A}\) and \(\overline{A}\) are two adjacent neighbors of A.

\textbf{Theorem 1.}\cite{c13} Consider a complex system that is virtually decomposed into subsystems. If all the decomposed subsystems are virtually stable in the sense of definition 1, then the entire system is stable. 

As can be seen from (\ref{equ8}) and (\ref{equ9}), 13 adaptation gains must be tuned and 13 lower-bound and 13 upper-bound parameters must be designated for a single subsystem, which for a system with \(n\) rigid bodies, they become 13\(n\) and 26\(n\). The aim of this paper is to reduce the amount of required parameters by replacing the projection function with the NAL adaptation function. This replacement eliminates all the aforementioned burdens by reducing 13\(n\) gains to 1 and eliminating 26\(n\) inertial parameter bounds. However, as can be seen from (\ref{equ6}), \(\theta_A \in \Re^{13}\) contains 13 parameters, whereas NAL requires \(\theta_A\) to be \(\Re^{10}\). Consequently, the next step is to reformulate the prevailing equations.

\subsection{Redesigned Dynamic Equations}

In this part, a new structure for the VDC scheme is proposed. This structure equips VDC with 10 unique representations of the rigid body in space with respect to frame \{A\} and puts it in a natural form, making it compatible with NAL. By defining \(\Bar{I}_A = I_A-m_A(^Ar_{AB}\times)^2\), we can rewrite (\ref{equ5}) as
\begin{equation}\label{equ14}
M_A = \begin{bmatrix}
m_A\, I_3 & -m_A\, (^Ar_{AB}\times) \\
m_A\, (^Ar_{AB}\times) & \Bar{I}_A
\end{bmatrix}\\
\end{equation}
\begin{equation*}
C_A = \begin{bmatrix}
m_A\, (^A\omega\times) & -m_A\, (^A\omega\times)\, (^Ar_{AB}\times) \\
m_A\, (^Ar_{AB}\times)\, (^A\omega\times) &  (^A\omega\times)\Bar{I}_A^A
\end{bmatrix}.
\end{equation*}
Details are presented in Appendix A. It must be mentioned that gravity vector does not need to be reformulated. Consequently, the linear-in-parameter form of new matrices can be written as
\begin{equation}\label{equ15}
    W_A\phi_A = M_A\frac{d}{dt}(^AV_r)+C_A(^AV_r)+G_A,
\end{equation}
where \(W_A \in \Re^{6\times10} \) is the new regressor matrix and \(\phi_A \in \Re^{10}\) is the 10-form inertial parameter vector. Details can be found in Appendix B. It is worth mentioning that the reqressor matrix represented in (\ref{B2}) has a more compact and simple form than the one in \cite{c13}. Now, by using (\ref{equ15}), it is possible to merge VDC and NAL.

\section{Natural Adaptation Law}

In the previous section, we derived a 10-form inertial parameter vector for the VDC context. This vector, \(\phi_A \in \Re^{10}\) (detailed in Appendix B), is a unique representation of a rigid body in space that contains mass and inertia represented in the frame \{A\}. However, to have such a unique representation just a subset of \(\Re^{10}\) can be assigned to \(\phi_A\). Such a subset must satisfy physical consistency conditions to be a unique representation of the rigid body. 

\textbf{Proposition 1.}\cite{c22} Consider a non-negative mass density function \(\mu: \Re^3 \rightarrow \Re^+_0\), such that
\begin{equation}\label{equ17}
    L_A = \int\begin{bmatrix}
        \textbf{x}_A \\ 
        1
    \end{bmatrix}
    \begin{bmatrix}
        \textbf{x}_A \\ 
        1
    \end{bmatrix}^T
    \mu(\textbf{x}_A)dV_A = 
    \begin{bmatrix}
        \Sigma_A & h_A \\
        h^T_A & m_A
    \end{bmatrix}
\end{equation}
where \(L_A \in S(4)\) is a \(4\times4\) symmetric matrix, \(\Sigma_A = \int\textbf{x}_A\textbf{x}^T_A\mu(\textbf{x}_A)dV_A \in S(3)\) is a second moment matrix, and \(S(n)\) is the space of \(n\times n\) real-symmetric matrices. The rigid body inertial parameter vector \(\phi_A\) is physically consistent if there exists a one-to-one linear map \(f:\Re^{10} \rightarrow S(4)\),

\begin{equation}\label{equ18}
    f(\phi_A)= L_A = \begin{bmatrix}
        0.5tr(\Bar{I}_A).\textbf{1}-\Bar{I}_A & h_A \\
        h^T_A & m_A
    \end{bmatrix}
\end{equation}
\begin{equation}\label{equ19}
    f^{-1}(\phi_A) = \phi_A(m_A,h_A,tr(\Sigma_A).\textbf{1}-\Sigma_A)
\end{equation}
such that \(L_A\) is positive-definite (i.e., \(L_A \succ 0\)). \(tr(.)\) is the trace operator and \(\Sigma_A = 0.5tr(\Bar{I}_A)-\Bar{I}_A\).

Based on proposition 1, the set of physically consistent inertial parameter vectors for a rigid body can be defined on manifold \(\mathcal{M}\) as

\begin{equation}\label{equ20}
\begin{split}
        \mathcal{M} = \{\phi_A \in \Re^{10}: f(\phi_A)\succ0\} \subset \Re^{10} \\
        = \{L_A \in S(4): L_A\succ0\}=\mathcal{P}(4)
\end{split}
\end{equation}
where \(\mathcal{P}(4)\) is the space of all real-symmetric, positive-definite matrices.

\subsection{Riemannian Geometry of \(\mathcal{P}(n)\) and Bregman Divergence}

\(T_P\mathcal{P}(n)\) is the tangent space for a given \(P\in \mathcal{P}(n)\). Therefore, Riemannian metric invariant under group action \(G*P = GPG^T\), where \(G \in GL(n)\) is any \(n\times n\) nonsingular matrix, for given \(P\in \mathcal{P}(n)\) and \(\mathfrak{F}_1,\mathfrak{F}_2 \in T_P\mathcal{P}(n)\) can be defined as

\begin{equation}\label{equ21}
    \langle \mathfrak{F}_1,\mathfrak{F}_2 \rangle_P = \frac{1}{2}tr(P^{-1}\mathfrak{F}_1P^{-1}\mathfrak{F}_2).
\end{equation}
Then, the geodesic distance between two arbitrary points \(L_1, L_2 \in \mathcal{P}(n)\) can be written in the sense of (\ref{equ21})
\begin{equation}\label{equ22}
    d_{\mathcal{P}{(4)}}(L_1,L_2) = (\sum_{i=1}^{n} (log(\lambda_i))^2){}^\frac{1}{2}
\end{equation}
where \(\lambda_i\) are the eigenvalues of \(L_1^{-1/2}L_2L_1^{-1/2}\). Thereafter, a distance metric on \(\mathcal{M}\) can be defined as
\begin{equation}\label{equ23}
    d_\mathcal{M}(\phi_A^1,\phi_A^2) = d_{\mathcal{P}{(4)}}(L_A^1,L_A^2).
\end{equation}

One other possible way to define distance metric is Bregman divergence. Let \(F:\ \Omega\rightarrow\Re\) be a continuous-differentiable, strictly convex function defined on a closed convex set \(\Omega\). The Bregman distance is defined as

\begin{equation}\label{equ24}
    D_F(\mathfrak{a},\mathfrak{b})=F(\mathfrak{a})-F(\mathfrak{b})-\langle \nabla F(\mathfrak{b}),\mathfrak{a}-\mathfrak{b}\rangle.
\end{equation}
By considering \(\Omega = \mathcal{P}(4)\) and assigning log-det function to \(F\) (i.e., \(F(P)=-log|P|\) for \(P \in \mathcal{P}(4)\)), the resulting Bregman divergence is,

\begin{equation}\label{equ25}
\begin{split}
    D_{F(\mathcal{P}(4))}(L_1\|L_2) = log\frac{|L_2|}{|L_1|}+tr(L_2^{-1}L_1)-4\\
    = \sum_{i=1}^{4}(-log(\lambda_i)+\lambda_i-1),
\end{split}
\end{equation}
where \(\lambda_i\) are the eigenvalues of \(L_2^{-1}L_1\). Now, by utilizing a one-to-one mapping \(f\) from \(\phi_A\) to \(L_A = f(\phi_A)\), a distance metric on \(\mathcal{M}\) can be expressed as

\begin{equation}\label{equ26}
    d_\mathcal{M}(\phi_A^1,\phi_A^2) = D_{F(\mathcal{P}(4))}(L_A^1\|L_A^2).
\end{equation}

\subsection{Extracted NAL}

In this part, the NAL function is embedded into VDC and proves that asymptotic stability is ensured. Consider a non-negative accompanying function \(\nu_T(t)\)

\begin{equation}\label{equ27}
    \nu_T(t) = \nu_P(t) + \nu_A(t)
\end{equation}
where \(\nu_P(t)\) is in the sense of (\ref{equ12}), and \(\nu_A(t)\) is the accompanying function of the adaptive term. By taking the derivative of (\ref{equ27}) and utilizing (\ref{equ15}) and the control law in the sense of (\ref{equ7}), the following is achieved:

\begin{equation}\label{equ28}
    \Dot{\nu}_T(t) \leq -X(t)QX(t)+p_{\underline{A}}-p_{\overline{A}}-\widetilde{\phi}_A^Ts(t) + \Dot{\nu}_A(t)
\end{equation}
where \(\widetilde{\phi}_A = \hat{\phi}_A-\phi_A\) and \(s(t)\) is defined in (\ref{equ10}). Most importantly, \(\Dot{\nu}_A(t)\) must possess the feature in the form of

\begin{equation}\label{equ29}
    \Dot{\nu}_A(t) = \widetilde{\phi}_A^T\mathcal{Z}(\Dot{\hat{\phi}},\hat{\phi}).
\end{equation}
The common quadratic function, which results in an adaptation function like the projection function, embodies such a feature. Another possible candidate for \(\nu_A(t)\) can be geodesic distance in the form of (\ref{equ22}). However, due to the nonlinearity of (\ref{equ22}), its time derivative can not be written in the form of (\ref{equ29}). Therefore, stability of the overall system cannot be guaranteed. Another choice is to consider \(\nu_A(t)\) as the Bregman divergence distance metric\cite{c21}:

\begin{equation}\label{equ30}
    \nu_A(t) = \gamma D_{F(\mathcal{P}(4))}(L_A\|\hat{L}_A)
\end{equation}
where \(\gamma>0\). From (\ref{equ25}) and according to the fact that \(-log(x)+x-1\geq0\), it can be concluded that (\ref{equ30}) is a valid accompanying function. Taking the derivative of (\ref{equ30}) leads to

\begin{equation}\label{equ31}
    \Dot{\nu}_A(t) = \gamma tr([\hat{L}_A^{-1}\Dot{\hat{L}}_A\hat{L}_A^{-1}]\widetilde{L}_A)
\end{equation}
where \(\widetilde{L}_A = \hat{L}_A-L_A\).

\begin{figure}[ht]
      \centering
      \includegraphics[scale = 0.4]{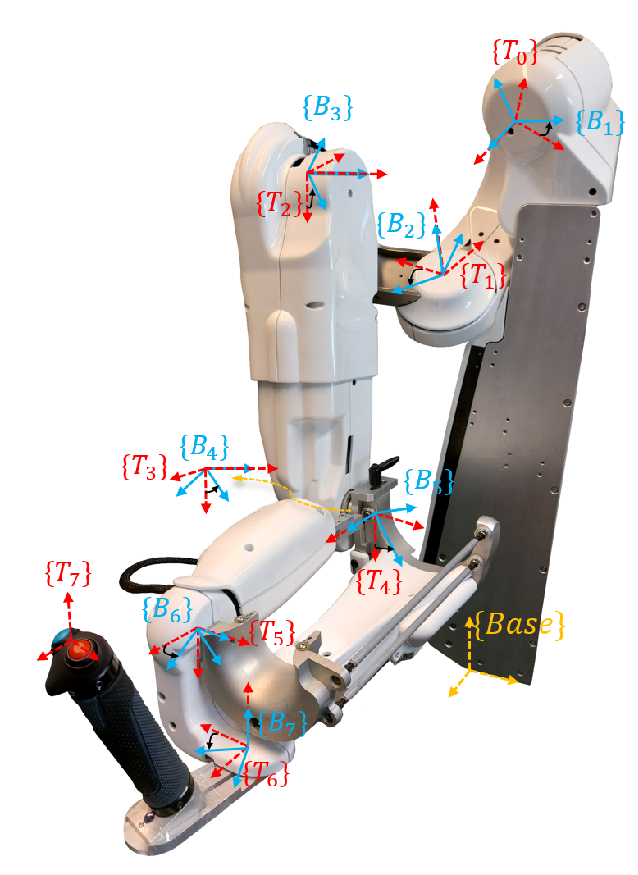}
      \caption{Commercial ABLE exoskeleton and assigned VDC frames}
     \label{figurelabel}
   \end{figure}

\textbf{Proposition 2.} Suppose that the adaptive VDC control law in the sense of (\ref{equ7}) results in a time derivative of the accompanying function in the form of (\ref{equ28}). Considering \(\nu_A(t)\) as defined in (\ref{equ30}) and its time derivative in (\ref{equ31}), the NAL is defined as\cite{c21}

\begin{equation}\label{equ32}
    \Dot{\hat{L}}_A = \frac{1}{\gamma} \hat{L}_AS\hat{L}_A
\end{equation}
and results in asymptotic stability of the system. \(S\) is a unique symmetric matrix that satisfies \(\widetilde{\phi}_A^Ts = tr(\widetilde{L}_A S)\).

\textbf{Proof.} Substituting (\ref{equ31}) in (\ref{equ28}) and using (\ref{equ32}) leads to
\begin{align*}
    \Dot{\nu}_T(t) \leq -X(t)QX(t)+p_{\underline{A}}-p_{\overline{A}}-\widetilde{\phi}_A^Ts(t) \\
    + \gamma tr([\hat{L}_A^{-1}\Dot{\hat{L}}_A\hat{L}_A^{-1}]\widetilde{L}_A)\\
    \leq -X(t)QX(t)+p_{\underline{A}}-p_{\overline{A}}-tr(\widetilde{L}_A S) \\
    + \gamma tr([\hat{L}_A^{-1}\Dot{\hat{L}}_A\hat{L}_A^{-1}]\widetilde{L}_A)\\
    \leq -X(t)QX(t)+p_{\underline{A}}-p_{\overline{A}} \qquad \qquad 
\end{align*}
which is in the sense of definition 1. Therefore, by using theorem 1, asymptotic stability of the whole system is established.

As demonstrated in (\ref{equ32}), there is only one tuning gain for the estimation of \(n\) rigid bodies in the absence of lower and upper limits for the inertial parameters vector. In addition, the NAL function ensures that all the estimated parameters will remain physically consistent by ascertaining that \(\hat{L}_A\) will remain positive-definite\cite{c21}.

\section{7 DOF Exoskeleton Control}

\begin{figure*}[ht]
     \centering
     \includegraphics[scale = 0.25]{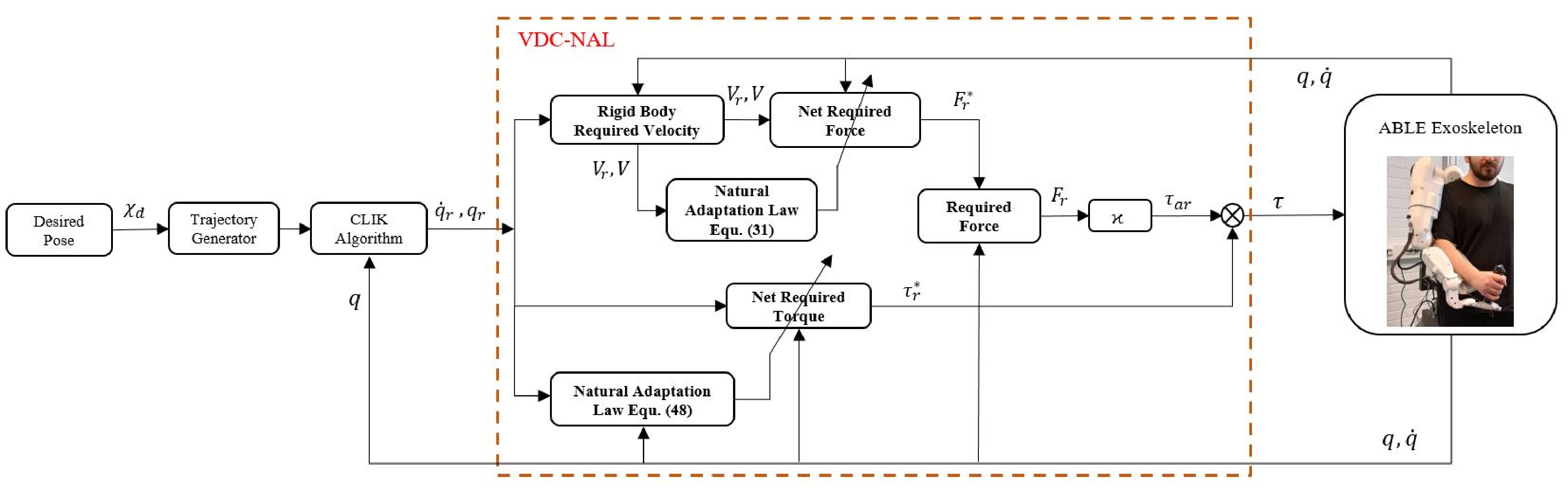}
     \caption{Block diagram of control scheme}
     \label{figurelabel}
\end{figure*}
In this section, the modified VDC is employed to control the 7-DoF commercial ABLE exoskeleton. Assigned VDC frames are demonstrated in Fig. 1. Using these frames, the entire robot configuration is decomposed into subsystems of joints and rigid bodies. It is assumed that joints have negligible mass. 

\subsection{Rigid Body Control Equations}
The velocity vector can be computed as

\begin{equation}\label{equ33}
     ^{B_i}V =\, ^{T_{i-1}}U_{B_i}^T\, ^{T_{i-1}}V + \varkappa_i \Dot{q}_i 
\end{equation}
\begin{equation}\label{equ34}
     ^{T_i}V =\, ^{B_i}U_{T_i}^T\, ^{B_i}V
\end{equation}
where \(i = 1 ... 7\), \(^{T_0}V=\, ^GV\) is the velocity vector of the ground, and \(^{T_7}V\) is the velocity vector of the end-effector, \(q_i\) and \(\Dot{q}_i\) are the joint angles and velocities, respectively, \(\varkappa_i = z_\tau\) for \(i = 1,2,3,4,6\), \(\varkappa_i = y_\tau\) for \(i = 7\), and \(\varkappa_i = x_\tau\) for \(i = 5\), which \(z_\tau = [0,0,0,0,0,1]^T\), \(y_\tau = [0,0,0,0,1,0]^T\), and \(x_\tau = [0,0,0,1,0,0]^T\). Moreover, \(^{B_i}U_{T_i}\) can be computed with (\ref{equ2}) by replacing \(B\) and \(A\) with \(T_i\) and \(B_i\), respectively. The required velocities also can be expressed as

\begin{equation}\label{equ35}
     ^{B_i}V_r =\, ^{T_{i-1}}U_{B_i}^T\, ^{T_{i-1}}V_r + \varkappa_i \Dot{q}_{ir}
\end{equation}
\begin{equation}\label{equ36}
     ^{T_i}V_r =\, ^{B_i}U_{T_i}^T\, ^{B_i}V_r
\end{equation}
where  \(^{T_0}V_r =\, ^GV_r\) is the required velocity of the ground, and \(^{T_7}V_r\) is the required velocity vector of the end-effector, and \(\Dot{q}_{ir}\) can be defined as follows:

\begin{equation}\label{equ37}
     \Dot{q}_{ir} = \textbf{J}^{-1}(\Dot{\chi}_d + \xi(\chi_d-\chi))
\end{equation}
where \(\textbf{J}\in \Re^{6\times7}\) is the Jacobian matrix, \(\Dot{\chi}_d \in \Re^6\) and \(\chi_d \in \Re^6\) are the desired velocity and pose of the end-effector in the Cartesian space, respectively, \(\xi \in \Re^{6\times6}\) is the diagonal, positive-definite matrix, and \(\chi \in \Re^6\) is the computed pose of the end-effector. Here, the defined end-effector's pose error is

\begin{equation}\label{equ38}
     e = \chi_d-\chi = \begin{bmatrix}
         e_p \\
         e_o
     \end{bmatrix}
\end{equation}
where \(e_p = p_d-p\) is the common position error with \(p = [x,y,z]^T\), and \(e_o\) is the quaternion-based orientation error, defined as

\begin{equation}\label{equ39}
     e_o = \eta(q)\epsilon_d-\eta_d\epsilon(q)-S(\epsilon_d)\epsilon(q)
\end{equation}
where \(\eta(q)\) and \(\epsilon(q)\) are unit quaternions computed from the rotation matrix, \(\eta_d\) and \(\epsilon_d\) are desired unit quaternions computed from the desired Euler angles of the end-effector with \(\alpha, \beta, \delta\) angles related to \(XYZ\) convention, and \(S(.)\) is a matrix operator. Now, the net force/moment vectors can be computed for each rigid body by replacing \(A\) with \(B_j\) in the (\ref{equ4}) for \(j = 7 ... 1\). Then, the force/moment vectors can be expressed as

\begin{equation}\label{equ40}
    ^{B_j}F =\, ^{B_j}U_{T_j}\, ^{T_j}F +\, ^{B_j}F^*
\end{equation}
\begin{equation}\label{equ41}
    ^{T_{j-1}}F =\, ^{T_{j-1}}U_{B_j}\, ^{B_j}F 
\end{equation}
with required terms as

\begin{equation}\label{equ42}
    ^{B_j}F_r =\, ^{B_j}U_{T_j}\, ^{T_j}F_r +\, ^{B_j}F_r^*
\end{equation}
\begin{equation}\label{equ43}
    ^{T_{j-1}}F_r =\, ^{T_{j-1}}U_{B_j}\, ^{B_j}F_r
\end{equation}
where \(^{T_7}F\) is the interaction force between the end-effector and environment, and \(^{T_0}F\) is the applied force from joint 1 to the ground. \(^{B_j}F_r^*\) is the required net force/moment vector that is defined in (\ref{equ7}).

\begin{equation}\label{equ44}
    ^{B_j}F_r^* = W_{B_j}\hat{\phi}_{B_j} + K_{B_j}(^{B_j}V_r-\,^{B_j}V).
\end{equation}
As introduced in the previous section, each of \(\hat{\phi}_{B_i}\) are updated by NAL (\ref{equ32}) by replacing \(A\) with \(B_i\) in (\ref{equ32}) with one adaptation gain \(\gamma\) for entire chain of rigid bodies.

\subsection{Virtual stability}

In this section, the virtual stability of the rigid body dynamics (\ref{equ15}) under the control action of (\ref{equ44}) is established. 

\textbf{Lemma 1.} If the accompanying function for \(i^{th}\) links with the control of (\ref{equ44}) and the adaptation law of (\ref{equ32}) is chosen as

\begin{equation}\label{equ45}
\begin{split}
\nu_{Ti}(t) = \frac{1}{2}(^{B_i}V_r-\,^{B_i}V)^TM_{B_i}(^{B_i}V_r-\,^{B_i}V) +\\ \frac{1}{2\gamma}\sum_{i=1}^{n} D_{F(\mathcal{P}(4))}(L_{B_i}\|\hat{L}_{B_i})
\end{split}
\end{equation}
it results in
\begin{equation}\label{equ46}
\begin{split}
\Dot{\nu}_{Ti}(t) = -(^{B_i}V_r-\,^{B_i}V)^T K_{B_i}(^{B_i}V_r-\,^{B_i}V) +\\
(^{B_i}V_r-\,^{B_i}V)^T(^{B_i}F_r^*-\,^{B_i}F^*).
\end{split}
\end{equation}
which shows the virtual stability of the link's dynamics in the sense of definition 1. 

\textbf{Proof.} Proof can be established by means of proposition 2 and \cite{c13}.

\subsection{Joint Control Equations}

The dynamic equations of joints in the absence of friction can be expressed as
\begin{equation}\label{equ47}
    I_{mi}\Ddot{q}_i=\tau^*_i = \tau_i-\tau_{ai}
\end{equation}
for \(i=1,...,7\), \(I_{mi}\) are the moment of inertia of the \(i^{th}\) joint, \(\tau^*_i\) denotes the net torque applied to the \(i^{th}\) joint, \(\tau_i\) is the control torque of the \(i^{th}\) joint, and \(\tau_{ai} = \varkappa_i^T\, ^{B_i}F\). The required net torque for each joint can be defined as
\begin{equation}\label{equ48}
    \tau^*_{ir} = W_{ai}\hat{\phi}_{ai}+k_{ai}(\Dot{q}_{ir}-\Dot{q}_{i})
\end{equation}
where \(k_{ai}>0\), \(W_{ai} = \varkappa_i^T\Ddot{q}_{ir}\), \(\hat{\phi}_{ai}\) are estimations of inertial parameter vectors for each motor with the update function of
\begin{equation}\label{equ49}
    \Dot{\hat{L}}_{ai} = \frac{1}{\gamma_a} \hat{L}_{ai}S_{ai}\hat{L}_{ai}
\end{equation}
where \(\hat{L}_{ai}\) is corresponding pseudo inertia matrix. Therefore, the control torque can be designed as
\begin{equation}\label{equ50}
    \tau_{i} = \tau^*_{ir} + \tau_{air}
\end{equation}
where \(\tau_{air} = \varkappa_i^T\, ^{B_i}F_r\).

\subsection{Virtual stability}

\textbf{Lemma 2.} If the accompanying function for the \(i^{th}\) joint with control action of (\ref{equ50}) and adaptation law of (\ref{equ49}) is chosen as
\begin{equation}\label{equ51}
\begin{split}
\nu_{a_i}(t) = \frac{1}{2}I_{mi}(\Dot{q}_{ir}-\Dot{q}_{i})^2 + \frac{1}{2\gamma_a}\sum_{i=1}^{n} D_F(L_{ai}\|\hat{L}_{ai})
\end{split}
\end{equation}
it results in
\begin{equation}\label{equ52}
\begin{split}
\Dot{\nu}_{ai}(t) = -k_{ai}(\Dot{q}_{ir}-\Dot{q}_{i})^2 +(\Dot{q}_{ir}-\Dot{q}_{i})(\tau^*_{ir}-\tau^*_{i}).
\end{split}
\end{equation}
which is virtually stable in the sense of definition 1. 

\textbf{Proof.} Proof is the same as lemma 1.

\textbf{Theorem 2.} If the accompanying function of the entire system is considered as
\begin{equation}\label{equ53}
    \nu(t) = \sum_{i=1}^{n} \nu_{a_i}(t) + \nu_{Ti}(t) 
\end{equation}
it follows,
\begin{equation}\label{equ54}
\begin{split}
   \Dot{\nu}(t) = -\sum_{i=1}^{n} ((^{B_i}V_r-\,^{B_i}V)^T K_{B_i}(^{B_i}V_r-\,^{B_i}V)\\
    + k_{ai}(\Dot{q}_{ir}-\Dot{q}_{i})^2).
\end{split}
\end{equation}
This shows that the entire system is asymptotically stable in the sense of theorem 1. 

\textbf{Proof. } The procedure is the same as Appendix C of \cite{c23}.

\section{Experimental Results}

In this section, the experimental results are presented to exhibit the performance of the new adaptive VDC method. The commercial upper-limb exoskeleton ABLE is utilized for the experiments. The haptic technology that this robot contains makes it a perfect choice for bilateral teleoperation, which is the goal of our future work. All high-level processes are accomplished in a host computer, and computed torque signals are transmitted to the robot using SIMULINK and the Virtuose interface, provided by Haption, with a sample time of 1 ms. Experiments are performed through the following scenario, as shown in Fig. 2. First, the desired Cartesian space targets are given to the \(5^{th}\) order trajectory generator \cite{c24}, which generates smooth acceleration trajectories in Cartesian space. Then, the required joint velocities are derived by employing the Closed-Loop Inverse Kinematic (CLIK) algorithm \cite{c25}, which is an online inverse kinematic solver. The computed required joint velocities are then fed to the proposed adaptive VDC method to compute the required net torques to execute Cartesian-space tasks.

To show that the new adaptive VDC has the same performance as the original one, simulation results related to the joint control of a 3-DoF spatial RRR robot utilizing both the original and proposed adaptive VDC are demonstrated in Fig. 3, where the solid lines are the tracking error of each joint with the projection function and the dashed lines correspond to the NAL function. The control objective is to track the desire \(sin(t)\) trajectory for each joint, starting from the zero initial condition. All the adaptation and control gains are considered the same. As can be seen, the performance of the proposed adaptive VDC is as good as the original one, and the tracking error is almost zero. The difference is that the NAL function requires only one gain for tuning, but the original adaptive VDC requires 39 adaptation gains and 78 upper and lower limits. This shows that applying the proposed adaptive VDC to high-DoF systems can be highly time-efficient.
\begin{figure}[ht]
      \centering
      \includegraphics[scale = 0.4]{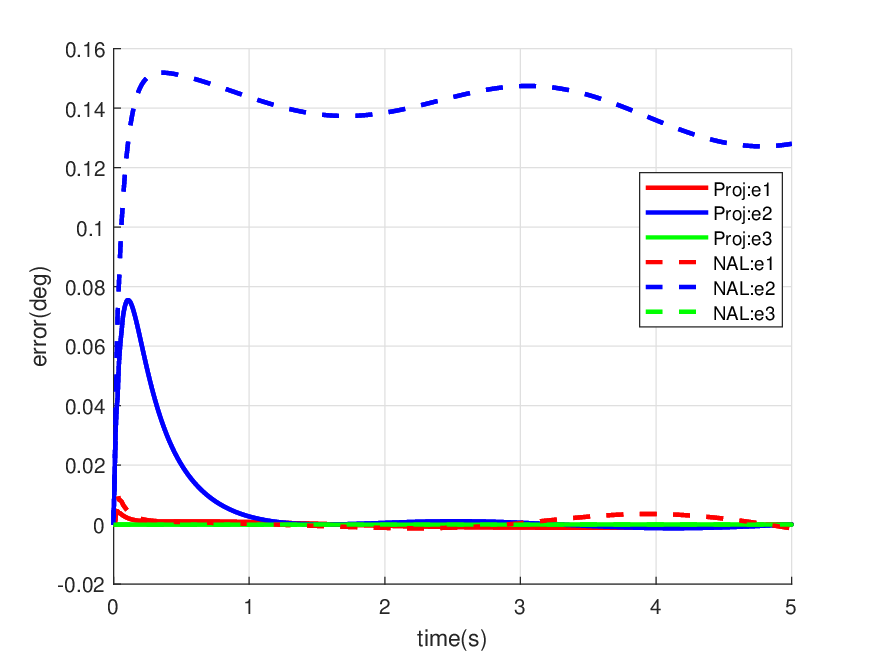}
      \caption{Joint tracking error for 3 DoF robot with Projection and NAL function}
      \label{figurelabel}
   \end{figure}
   
In the following part, the experimental results of the pose control of the 7-DoF redundant exoskeleton are presented. It must be mentioned that to activate joint motors, the human operator must grasp the handle of the exoskeleton, which, as can be seen in Fig. 1, is the end-effector. Therefore, the intention force of the operator, which can be considered an unknown time-varying disturbance, affects the performance of the controller in reaching the desired pose. Moreover, the utilized exoskeleton is a redundant robot that can have different inverse kinematic solutions each time the robot performs a task, especially for different initial values of robot configuration. To have reliable results for the performance of the proposed controller, the experiment is repeated 6 times, and the means and standard deviations for position and orientation errors are computed. The utilized control and adaptation gains are as follows: \(\xi=25I_{6\times6},\, K_{B_i}=1.2,\, k_{a_i}=0.1,\, \gamma=\gamma_a=10\). The initial condition for the adaptation function of (31) is chosen in a way that satisfies the physical consistency condition and \(^GV=\,^GV_r=\,^{T_7}F=\,^{T_7}F_r=\,\textbf{0}_{6\times1}\). DH parameters and the Jacobian matrix are derived based on the convention explained in \cite{c26}. 

Fig. 4(a) demonstrates the mean and standard deviation of the position errors of the end-effector in each direction, and Fig. 4(b) shows a sample of desired path tracking. As can be seen in Fig. 4, the proposed adaptive VDC has excellent performance in tracking the desired path in Cartesian space. The results demonstrate that the adaptive VDC is robust against unknown time-varying disturbances and stably maintains the system throughout the working time. The average Root Mean Square Error (RMSE) of position errors for all experiments is \(11.3\pm1 mm\). Moreover, Fig. 5 shows the mean and standard deviation of the quaternion orientation errors of the end-effector computed with (\ref{equ39}). It can be seen that in the presence of unknown external time-varying disturbances, which are directly applied to the end-effector, adaptive VDC exhibits perfect tracking of the desired orientation. The average RMSE of orientation errors for all experiments is \(1.06\pm0.25 deg\).

\begin{figure}[ht]
      \centering
      \includegraphics[scale = 0.4]{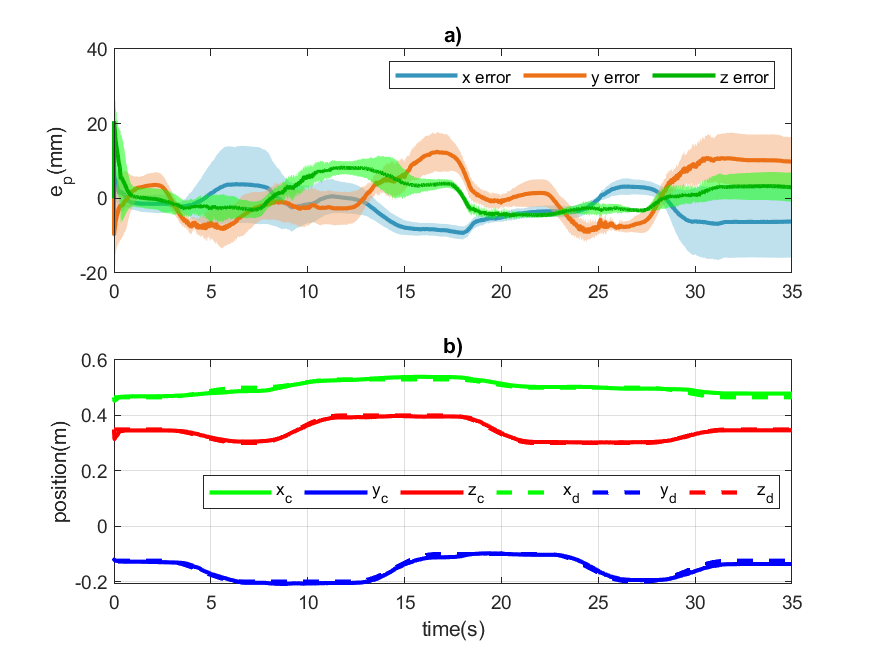}
      \caption{End-effector position error a) The mean and standard deviation of position errors in 6 experiments b) Cartesian path following}
      \label{figurelabel}
   \end{figure}

\begin{figure}[ht]
      \centering
      \includegraphics[scale = 0.4]{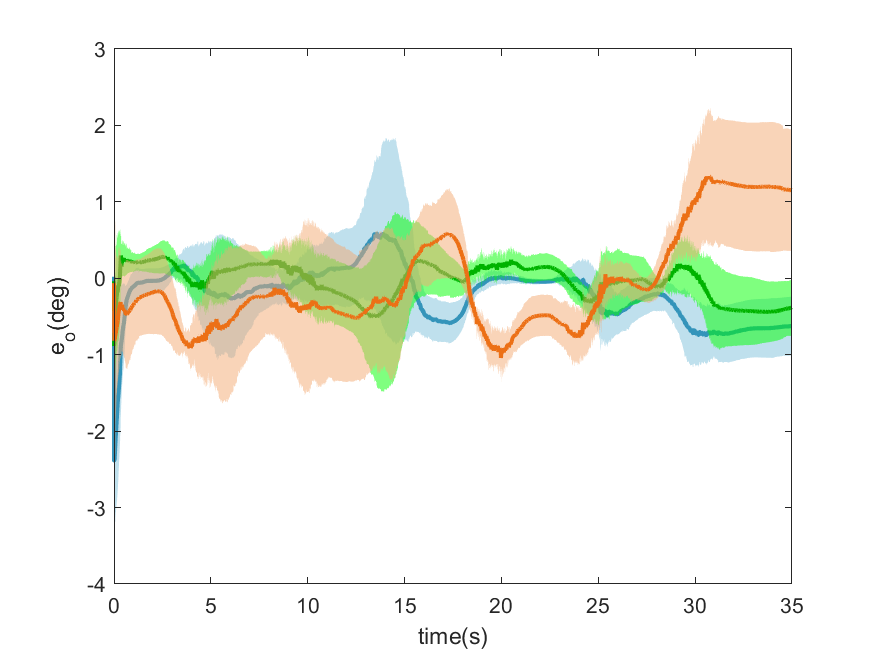}
      \caption{End-effector orientation error}
      \label{figurelabel}
   \end{figure}

\section{CONCLUSIONS}
In this paper, we proposed a new adaptive VDC approach that uses NAL to estimate unknown inertial parameters. The original adaptive VDC requires tuning of 13\(n\) adaptation gains and assigning 26\(n\) upper and lower parameter limits, which makes the process excessively tedious. However, the proposed adaptive VDC removes all the mentioned burdens and requires only one adaptation gain for all the rigid bodies. First, we reformulated the dynamic equations of the VDC context to put them into 10 unique inertial parameter vectors, making them compatible with the NAL function. Then, it was proved that employing the NAL function as the adaptation function for adaptive VDC culminates in asymptotic stability. Finally, the experimental results of applying the new adaptive VDC to the redundant upper-limb exoskeleton demonstrated the excellent performance of the VDC approach.

Our future work target is the control problem of two master exoskeleton robots that command heavy-duty hydraulic mobile manipulators as slaves for handling common objects in human-like manipulation scenarios with high scaling. As we showed in our earlier work \cite{c3}, the high motion/force scaling must be tackled in the controller design and stability analysis, which makes analysis much more complicated. Therefore, a novel adaptive VDC scheme developed in this paper will greatly lessen the implementation burden of this highly complex MMMS system.

\section*{APPENDIX}

\subsection{Restructured Dynamic Equations}

To rewrite the \(C_A\) matrix, it is enough to rewrite \(C_A(2,2)\). By expanding \(C_A(2,2)\,^A\omega\), we get to

\renewcommand{\theequation}{A.\arabic{equation}}
\setcounter{equation}{0}
\begin{equation}\label{A1}
\begin{split}
    C_A(2,2)\,^A\omega = (^A\omega\times)I_A(^A\omega)+I_A(^A\omega\times)(^A\omega)\\
    -m_A(^Ar_{AB}\times)(^A\omega\times)(^Ar_{AB}\times)(^A\omega)
\end{split}\qquad
\end{equation}
where it is obvious that \(I_A(^A\omega\times)(^A\omega)\) is always zero. By using the fact that \(a\times b = -b\times a\) along with
\begin{equation*}
\begin{split}
m_A(^Ar_{AB}\times)(^A\omega\times)(^Ar_{AB}\times)(^A\omega) = m_A\,^A\omega\times\,\\(^Ar_{AB}\times)^2\,^A\omega
\end{split}
\end{equation*}
(\ref{A1}) can be reformulated as
\begin{equation*}
    C_A(2,2)\,^A\omega = (^A\omega\times)\Bar{I}_A\,^A\omega.
\end{equation*}
Therefore, the \(C_A\) matrix can be written in the form of (\ref{equ14}).

\subsection{Deriving \(W_A\) and \(\phi_A\)}

By completing the multiplications on the right-hand side of (\ref{equ15}), one can get,
\renewcommand{\theequation}{B.\arabic{equation}}
\setcounter{equation}{0}
\begin{equation}\label{B1}
\!\begin{aligned}
&
    \left[\begin{matrix}
        m_A\,^A(\Dot{v}_r)-m_A\,(^Ar_{AB}\times)\,^A\Dot{\omega}_r+m_A\,(^A\omega_r\times)\,\\
        m_A\,(^Ar_{AB}\times)\,^A(\Dot{v}_r)+\Bar{I}\,^A\Dot{\omega}_r+m_A\,(^Ar_{AB}\times)\,(^A\omega_r\times)\,^Av_r
    \end{matrix}\right.\\
& \quad
\left.\begin{matrix}
-m_A\,(^A\omega_r\times)\,(^Ar_{AB}\times)\,^A\omega_r+m_A\,^AR_{I}g\\
+(^A\omega_r\times)\,\Bar{I}\,^A\omega_r+m_A\,(^Ar_{AB}\times)\,^AR_{I}g
\end{matrix}\right] =\, ^AF_r^*.
\end{aligned} \,
\end{equation}
Then, by defining
\begin{equation*}
    vecI = [\Bar{I}_{xx},\Bar{I}_{yy},\Bar{I}_{zz},\Bar{I}_{xy},\Bar{I}_{yz},\Bar{I}_{xz}]^T
\end{equation*}
\begin{equation*}
    (.^A\omega) =\begin{bmatrix}
         \omega_x & 0 & 0 & \omega_y & 0 & \omega_z \\
         0 & \omega_y & 0 & \omega_x & \omega_z & 0 \\
         0 & 0 & \omega_z & 0 & \omega_y & \omega_x
    \end{bmatrix}
\end{equation*}
and obtaining \(a\times b = -b\times a\), (\ref{B1}) can be rewritten as
\begin{equation*}
    W_A(\,^A\Dot{v}_r,\,^A\Dot{\omega}_r,\,^Av_r,\,^A\omega_r,\,^AR_Ig)\phi_A=\, ^AF_r^*
\end{equation*}
where
\begin{equation}\label{B2}
\begin{aligned}
W_A &= 
    \left[\begin{matrix}
        ^A\Dot{v}_r+(^A\omega_r\times)\,^Av_r+^AR_{I}g \\ 
        \textbf{0}_{3\times 1} \\ 
    \end{matrix}\right.\\
& \qquad \qquad
\left.\begin{matrix}
    ^A\Dot{\omega}_r+(^A\omega_r\times)^2 \\
    -((^A(\Dot{v}_r)\times)+(^A\Bar{\omega v}\times)+(^AR_Ig\times))
    \end{matrix}\right.\\
& \qquad \qquad \qquad \qquad \quad
\left.\begin{matrix}
    \textbf{0}_{3\times 6}\\
    (.^A\Dot{\omega}_r)+(^A\omega_r\times)(.^A\omega_r)
\end{matrix}\right]
\end{aligned}
\end{equation}
and \(\phi_A = [m_A, h_A^T, vecI^T]^T\), \(h = m_A\,^Ar_{AB}\), and \(^A\Bar{\omega v} = (^A\omega_r\times)\,^Av_r\).

\end{document}